\documentclass[letterpaper]{article} 
\usepackage{aaai2026}  
\usepackage{times}  
\usepackage{helvet}  
\usepackage{courier}  
\usepackage[hyphens]{url}  
\usepackage{graphicx} 
\usepackage{natbib}  
\usepackage{caption} 
\usepackage{subcaption}
\usepackage{amssymb}
\usepackage{amsmath}
\usepackage{multirow}
\usepackage{booktabs}
\usepackage{makecell}
\usepackage{pifont}

\usepackage{color, xspace}

\newcommand{\name}{HyperD\xspace} 

\title{HyperD: Hybrid Periodicity Decoupling Framework for Traffic Forecasting
}

\author{
Minlan Shao\textsuperscript{\rm 1}, 
Zijian Zhang\textsuperscript{\rm 2,}\thanks{Corresponding authors: Zijian Zhang and Xin Wang.}, 
Yili Wang\textsuperscript{\rm 1}, 
Yiwei Dai\textsuperscript{\rm 1}, Xu Shen\textsuperscript{\rm 1}, 
Xin Wang\textsuperscript{\rm 1,}\footnotemark[1]}

\affiliations{
\textsuperscript{\rm 1} School of Artificial Intelligence, Jilin University, Changchun, China\\ \textsuperscript{\rm 2} College of Computer Science and Technology, Jilin University, Changchun, China\\
\{shaoml24, wangyl21, daiyw23, shenxu23\}@mails.jlu.edu.cn, 
\{zhangzijian, xinwang\}@jlu.edu.cn
}

\begin{document}
\maketitle

\begin{abstract}

Accurate traffic forecasting plays a vital role in intelligent transportation systems, enabling applications such as congestion control, route planning, and urban mobility optimization. However, traffic forecasting remains challenging due to two key factors: (1) complex spatial dependencies arising from dynamic interactions between road segments and traffic sensors across the network, and (2) the coexistence of multi-scale periodic patterns (e.g., daily and weekly periodic patterns driven by human routines) with irregular fluctuations caused by unpredictable events (e.g., accidents or weather disruptions).
To tackle these challenges, we propose \textbf{HyperD} (Hybrid Periodic Decoupling), a novel framework that decouples traffic data into \textbf{periodic} and \textbf{residual components}. The periodic component is handled by the \textbf{Hybrid Periodic Representation Module}, which extracts fine-grained daily and weekly patterns using learnable periodic embeddings and spatial-temporal attention. The residual component, which captures non-periodic, high-frequency fluctuations, is modeled by the \textbf{Frequency-Aware Residual Representation Module}, leveraging complex-valued MLP in frequency domain. To enforce semantic separation between the two components, we further introduce a \textbf{Dual-View Alignment Loss}, which aligns low-frequency information with the periodic branch and high-frequency information with the residual branch. Extensive experiments on four real-world traffic datasets demonstrate that HyperD achieves state-of-the-art prediction accuracy, while offering superior robustness under disturbances and improved computational efficiency compared to existing methods.
\end{abstract}

\begin{links}
    \link{Code}{https://github.com/ll121202/HyperD}
\end{links}

\section{Introduction}

The rapid development of urbanization and sensing technologies drives abundant spatial-temporal data across domains such as traffic systems~\cite{liu2024spatial1, zhou2020foresee, ma2023rethinking}, power grids~\cite{li2025causal}, and environmental monitoring~\cite{liu2025spatiotemporal}. Among these, traffic forecasting is a core task in intelligent transportation systems, as accurate flow prediction is essential for understanding and managing urban mobility dynamics~\cite{xia2025fairtp, zhou2025coms2t}. It  requires traffic forecasting models to well capture rich spatial dependencies and diverse temporal patterns inherent in real-world road networks.

\begin{figure}[t!]
    \centering
    \includegraphics[width=0.49\textwidth]{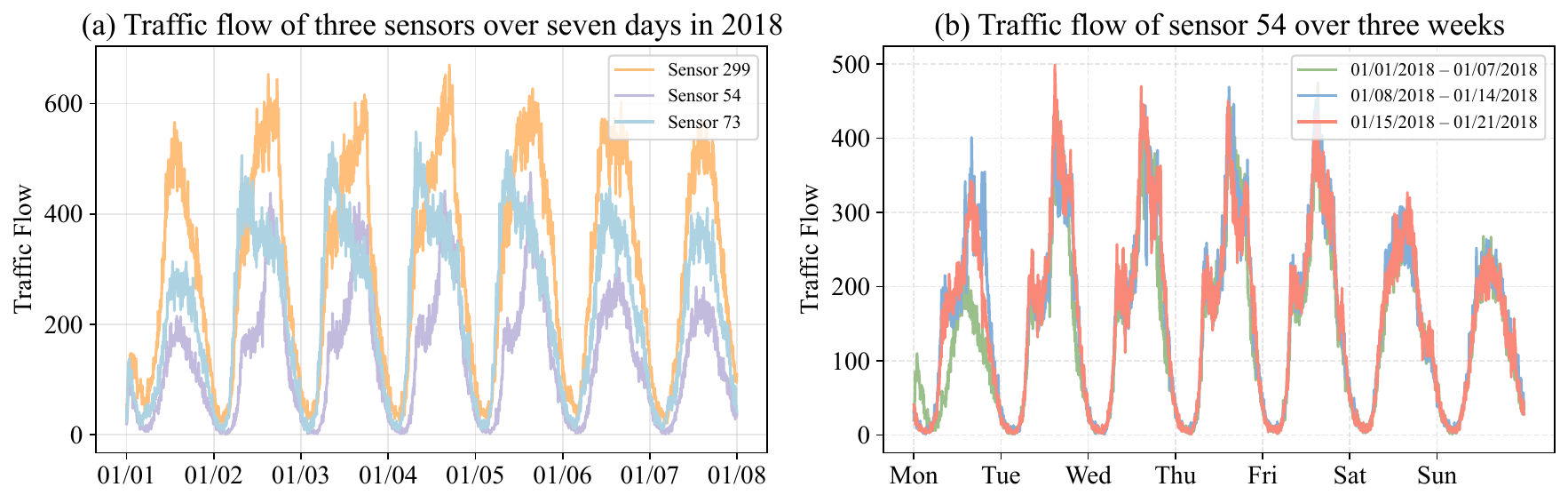} 
    \caption{Illustration of traffic flow in the PEMS04 dataset. (a) Traffic flow for sensors 54, 73, and 299 over one week. (b) Traffic flow of sensor 54 over three consecutive weeks. Both images demonstrate the strong presence of periodic patterns in traffic data.
}
\label{fig:1}
\end{figure}

Driven by consistent human routines and recurring activity patterns, periodicity plays a fundamental role in the temporal pattern of traffic data
~\cite{zhang2017deep, yu2019citywide}.
As illustrated in Figure~\ref{fig:1}, traffic flow curves often repeat in consistent patterns across days and weeks, while also showing notable variation across different sensor locations. This indicates that \textbf{periodicity is not only prominent but also spatially diverse}, making it a critical predictive signal. As a result, many existing methods attempt to leverage these patterns through trend-seasonality decomposition ~\cite{yuandiffusion, wang2024towards, cao2025spatiotemporal}, which separates the data into a trend component and a seasonal component. However, these methods suffer from two key limitations: \ding{192} \textbf{Implicit Modeling of Periodicity} — Periodic information is split across trend and seasonal components, overlooking its unified structure across time scales and spatial locations. This fragmentation hinders direct modeling of coherent periodic patterns.
\ding{193} \textbf{Inaccurate Decomposition} — Traditional decomposition methods are non-learnable and inefficient, often resulting in misalignment between trend and seasonal components, and poor performance in capturing complex temporal patterns~\cite{yu2024revitalizing, kim2025comprehensive}. 

To move beyond implicit periodic modeling, recent methods like CycleNet~\cite{cyclenet} make initial strides toward explicit periodic modeling. However, these approaches can hardly tackle traffic forecasting because they fail to model interactions between spatial locations and temporal dynamics. Additionally, they consider only a single periodic scale, leaving rich multi-scale periodic patterns underutilized. Addressing these limitations requires a flexible framework that can capture both multi-scale periodic patterns and fine-grained spatial-temporal interactions. Furthermore, periodic patterns in traffic data are often overlapping and spatially heterogeneous, while irregular fluctuations challenge conventional modeling approaches.

To cope with limitations above, we propose \name, a \underline{\textbf{Hy}}brid \underline{\textbf{Per}}iodic \underline{\textbf{D}}ecoupling framework that decouples traffic data into \textbf{periodic and residual components}, each modeled by tailored mechanisms. 
(1) For implicit modeling of periodicity, the \textbf{\textit{Hybrid Periodic Representation Module}} uses multiple learnable embeddings and a spatial-temporal attention mechanism to model multi-scale periodic patterns. The learnable embeddings are leveraged according to the timestamps of traffic states and maintain multi-scale periodic patterns explicitly,
addressing the challenge of overlapping periodic patterns that existing methods often fail to resolve. 
(2) For inaccurate decomposition, \name uses the \textbf{\textit{Frequency-Aware Residual Representation Module}} to model residual dynamics, and further introduce the \textbf{\textit{Dual-View Alignment Loss}} to ensure separation between the two branches. Specifically, this loss aligns low-frequency information with the periodic branch and high-frequency information with the residual branch, encouraging each component to focus on its respective frequency band for more precise decoupling. This promotes divergence between periodic and residual components, enabling HyperD to capture both \textbf{multi-scale periodic patterns} and \textbf{irregular fluctuations}.
Our main contributions are as follows:

\begin{itemize}
\item HyperD is proposed as a novel framework that decouples traffic data into periodic and residual components, addressing multi-scale periodic patterns and irregular fluctuations for more accurate forecasting.
\item A Hybrid Periodic Representation Module captures multi-scale periodic patterns using learnable embeddings and spatial-temporal attention mechanism, while a Frequency-Aware Residual Representation Module models non-periodic fluctuations through a spatial-temporal frequency encoder.
\item The Dual-View Alignment Loss ensures effective decoupling between periodic and residual components, preventing semantic redundancy. 
Extensive experiments demonstrate that HyperD outperforms existing methods in accuracy, robustness, and computational efficiency.
\end{itemize}
\section{Related Work}
\subsection{Spatial-Temporal Forecasting}
Spatial-temporal graph neural networks (STGNNs)~\cite{zhang2023mlpst, miao2024less, liu2024mvcar, wang2024unifying} have become a dominant approach in traffic forecasting, as they jointly model spatial correlations and temporal dynamics. 
The integration of LLMs with spatial-temporal models has also shown strong potential for multivariate time-series forecasting~\cite{timecma2025liu, zhang2023promptst, liu2025towards, shen2025understanding}.
For spatial modeling, recent approaches typically employ graph neural networks~\cite{wang2025adagcl+, GRASS, FuDHL0C25, shen2024graph, he2025mamba, wang2024goodat, shao2025enhanced}, to encode spatial correlations among sensors. Representative examples include STGCN~\cite{yu2018spatio} and DCRNN~\cite{li2018diffusion}, which build on fixed road network structures, while GWNet~\cite{wu2019graph} and AGCRN~\cite{bai2020adaptive} further introduce adaptive graphs to capture time-varying spatial dependencies.

For temporal modeling, various architectures have been explored. RNN-based models~\cite{jiang2023spatio} capture short-term sequential patterns, while TCNs~\cite{wu2019graph, fang2021spatial} employ dilated convolutions to model longer temporal dependencies. More recently, Transformer-based models~\cite{jiang2023pdformer, gao2024spatial} use attention mechanisms to capture complex temporal dynamics.

\subsection{Spatial-Temporal Decoupling Methods}
To handle the complexity of traffic dynamics, some recent methods attempt to decouple spatial-temporal data into interpretable components.
D\textsuperscript{2}STGNN~\cite{shao2022decoupled} decouples traffic signals into diffusion and inherent components to model different propagation patterns.
STWave~\cite{fang2023spatio} disentangles representations of traffic time series, separating complex traffic dynamics into stable trends and fluctuating events. 
STDN~\cite{cao2025spatiotemporal} decomposes traffic flow into the trend-cyclical and seasonal components in view of spatial-temporal embeddings.


\begin{figure*}[t!]
  \centering
\includegraphics[width=0.9\textwidth]{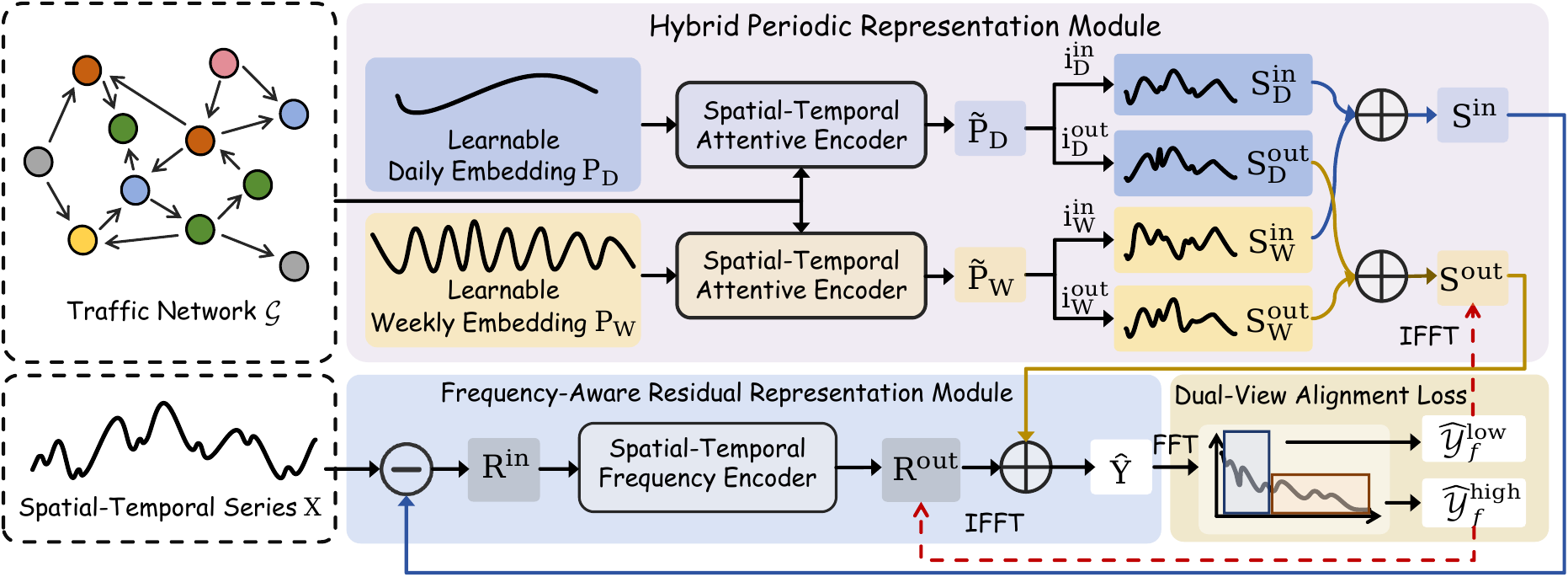} 
  \caption{Overview of \name, which comprises three main components: (1) The Hybrid Periodic Representation Module encodes daily and weekly embeddings using the Spatial-Temporal Attentive Encoder and generates hybrid periodic patterns.
  (2) The Frequency-Aware Residual Representation Module encodes the residual using a Spatial-Temporal Frequency Encoder and combines it with the periodic component to yield the final prediction.
  (3) The Dual-View Alignment Loss separates the prediction into low- and high-frequency parts, which are then aligned with the periodic and residual branches, respectively.}
  \label{fig:framework} 
\end{figure*}


\section{Methodology}
\textbf{Problem Definition.} 
We represent the traffic network as a directed graph $\mathcal{G} = \{\mathcal{V}, \mathcal{E}, \mathbf{A} \}$, where $\mathcal{V}$ is a set of $N$ nodes,  corresponding to the traffic sensors deployed on the road network. $\mathcal{E}$ denotes the set of edges indicating the connectivity between sensors. $\mathbf{A} \in \mathbb{R}^{N \times N}$ is the adjacency matrix that describes whether a connection exists between nodes.


Given a historical traffic time series
$\mathbf{X} = \{\boldsymbol{x}_1, \boldsymbol{x}_2, \dots, \boldsymbol{x}_{T_1}\} \in \mathbb{R}^{T_1 \times N}$, where $\boldsymbol{x}_t \in \mathbb{R}^N$ represents the observation at time $t$ across $N$ nodes, our goal is to predict the future traffic states over the next $T_2$ time steps. We denote the predicted time series as $\hat{\mathbf{Y}} = \{ \hat{\boldsymbol{y}}_{T_1+1}, \hat{\boldsymbol{y}}_{T_1+2}, \ldots, \hat{\boldsymbol{y}}_{T_1+T_2}\}\in \mathbb{R}^{T_2 \times N}$, which is obtained by:
\begin{equation}
    \mathbf{X}, \mathcal{G} \xrightarrow{f_\theta} \hat{\mathbf{Y}},
\end{equation}
where $f_\theta$ is a mapping function parameterized by $\theta$.

\subsection{Hybrid Periodic Representation Module}
The overall framework of \name is shown in Figure~\ref{fig:framework}.
To effectively capture multi-scale periodic patterns in traffic data, we explicitly model periodicity using learnable embeddings. This allows HyperD to leverage prior knowledge of daily and weekly temporal structures, which are among the most dominant patterns in traffic data. This approach can also be easily extended to other temporal patterns, such as hourly or monthly. 
The Hybrid Periodic Representation Module consists of three main components: Learnable Daily Embeddings (LDE) and Learnable Weekly Embeddings (LWE), Spatial-Temporal Attentive Encoder (STAE), and Hybrid Periodic Pattern.

\subsubsection{Learnable Daily and Weekly Embedding}
The goal of LDE and LWE is to explicitly capture the dominant daily and weekly periodic patterns in traffic data, which are essential for accurate traffic forecasting. To achieve this, we introduce two learnable embeddings:  $\mathbf{P}_D \in \mathbb{R}^{L_D \times N}$ for daily patterns and $\mathbf{P}_W \in \mathbb{R}^{L_W \times N}$ for weekly patterns, where $L_D$ and $L_W$ are the respective period lengths. These embeddings allow the model to effectively represent periodic patterns at multiple temporal scales.

The period lengths are determined by data sampling frequency. For instance, with a 5-minute sampling rate, the daily period length is $L_D = 288$ and the weekly period length is $L_W = 2016$. To accelerate the convergence and incorporate temporal prior knowledge, we initialize these embeddings using a statistical prior initialization strategy. We normalize the training data and compute the mean values for each node at every time step within the day and week. These mean values are used to initialize the embeddings, providing a reliable starting point for learning periodic patterns.

As a core component of the Hybrid Periodic Representation Module, these learnable embeddings captures the primary periodic patterns in the traffic data. 
To refine these embeddings and capture long-range dependencies, we introduce the STAE, which combines spatial correlations and temporal dynamics through attention mechanisms.



\subsubsection{Spatial-Temporal Attentive Encoder}
To model interactions among nodes and capture long-range dependencies, we refine the learned periodic embeddings using the STAE. It combines GCNs~\cite{kipf2017semi} with spatial and temporal self-attention mechanisms to integrate both spatial correlations and temporal dynamics. 
This encoder enhances the embeddings, ensuring that both local and global dependencies are effectively captured.

Given a periodic embedding $\mathbf{P} \in \mathbb{R}^{L \times N}$, where $L$ denotes the period length (e.g., $L_D$ for daily, $L_W$ for weekly), we first apply GCN to incorporate spatial correlations:
\begin{equation}
    \mathbf{H} = \sigma(\hat{\mathbf{A}} \mathbf{P} \mathbf{W}), 
\end{equation}
where $\hat{\mathbf{A}} \in \mathbb{R}^{N \times N}$ is the normalized Laplacian matrix, $\mathbf{W} \in \mathbb{R}^{N \times N}$ is the trainable weight matrix, and $\sigma$ denotes the ReLU activation function. 

To further capture long-range dependencies within the periodic patterns, we apply the self-attention mechanism~\cite{vaswani2017attention} along both temporal and spatial dimensions:
\begin{equation}
    \begin{split}
        \mathbf{Q} = \mathbf{H} \mathbf{W}^q \quad \mathbf{K} &= \mathbf{H} \mathbf{W}^k \quad \mathbf{V} = \mathbf{H} \mathbf{W}^v \\
        \mathrm{Attention}(\mathbf{H}) &= \mathrm{softmax} \left( \frac{\mathbf{Q} \mathbf{K}^\top}{\sqrt{d}} \right)\mathbf{V},
    \end{split}
\end{equation}
where $\mathbf{H}$ is the input feature matrix, and $\mathbf{W}^q, \mathbf{W}^k, \mathbf{W}^v$ are the trainable projection matrices. 

We perform temporal self-attention by treating each time step as a query, with trainable projections $\mathbf{W}^q_t, \mathbf{W}^k_t, \mathbf{W}^v_t \in \mathbb{R}^{N \times N}$. We perform spatial self-attention by transposing $\mathbf{H}$ and treating each node as a query, with trainable projections $\mathbf{W}^q_s, \mathbf{W}^k_s, \mathbf{W}^v_s \in \mathbb{R}^{L \times L}$. Let $\mathbf{H}_t = \mathrm{Attention}_t(\mathbf{H})$ and $\mathbf{H}_s = \mathrm{Attention}_s(\mathbf{H}^\top)^\top$ denote the outputs of temporal and spatial self-attention, respectively. The two outputs are concatenated and fused through a linear projection:
\begin{equation}
    \tilde{\mathbf{P}} = \mathbf{W}_o[\mathbf{H}_t; \mathbf{H}_s],
\end{equation}
where $\tilde{\mathbf{P}} \in \mathbb{R}^{L \times N}$, $\mathbf{W}_o \in \mathbb{R}^{2N*N}$ is the trainable weight matrix.
The same operations are independently applied to the daily and weekly embeddings, producing $\tilde{\mathbf{P}}_D \in \mathbb{R}^{L_D \times N}$ and $\tilde{\mathbf{P}}_W \in \mathbb{R}^{L_W \times N}$, respectively. As the operations are architecturally identical, we present a unified formulation above for clarity.
The refined embeddings, $\tilde{\mathbf{P}}_D$ and $\tilde{\mathbf{P}}_W$, are integrated to form the Hybrid Periodic Pattern. 

\subsubsection{Hybrid Periodic Pattern}
To integrate daily and weekly periodic patterns, we construct a hybrid periodic pattern by aggregating the corresponding segments from the daily and weekly embeddings.
For each time step, we use two temporal metadata: the \textbf{time of day} and the \textbf{day of week}, to compute the indices in the daily and weekly embeddings. The \textbf{daily index} is given by $\mathbf{i}_D = \text{time of day}$. The \textbf{weekly index} is computed as $\mathbf{i}_W = \text{time of day} + \text{day of week} \times L_D$. 
Given the input time series $\mathbf{X} \in \mathbb{R}^{T_1 \times N}$, we retrieve the corresponding segments from the daily and weekly embeddings, $\tilde{\mathbf{P}}_D$ and $\tilde{\mathbf{P}}_W$, using the indices $\mathbf{i}_D^{\mathrm{in}}$ and $\mathbf{i}_W^{\mathrm{in}}$. These segments are denoted as:
\begin{equation}
    \mathbf{S}^{\mathrm{in}}_D(t) = \tilde{\mathbf{P}}_D[
    \mathbf{i}^{\mathrm{in}}_D(t), :] \quad 
    \mathbf{S}^{\mathrm{in}}_W(t) = \tilde{\mathbf{P}}_W[\mathbf{i}^{\mathrm{in}}_W(t), :], 
\end{equation}
where $t \in \{1, \dots, T_1\}$ represents the time steps in the historical time series. These two segments are then aggregated to form the hybrid periodic pattern for the historical time series: $\mathbf{S}^{\mathrm{in}} = \mathbf{S}_D^{\mathrm{in}} + \mathbf{S}_W^{\mathrm{in}}$.

Similarly, for the predicted time series $\hat{\mathbf{Y}}$, we retrieve the corresponding daily and weekly segments for each time steps $t \in \{T_1+1, \dots, T_1+T_2\}$:
\begin{equation}
    \mathbf{S}^{\mathrm{out}}_D(t) = \tilde{\mathbf{P}}_D[
    \mathbf{i}^{\mathrm{out}}_D(t), :] \quad 
    \mathbf{S}^{\mathrm{out}}_W(t) = \tilde{\mathbf{P}}_W[\mathbf{i}^{\mathrm{out}}_W(t), :].
\end{equation}
They are aggregated to form the hybrid periodic pattern for the predicted time series: $\mathbf{S}^{\mathrm{out}} = \mathbf{S}_D^{\mathrm{out}} + \mathbf{S}_W^{\mathrm{out}}$.

The Hybrid Periodic Pattern captures the primary periodic dynamics in the data, setting a solid foundation for forecasting. However, traffic data also contains irregular fluctuations that cannot be captured by periodic patterns alone. To address this, we introduce the \textbf{Frequency-Aware Residual Representation Module}.

\subsection{Frequency-Aware Residual Representation Module}

While daily and weekly embeddings capture stable periodic patterns, traffic data also contains irregular fluctuations and disruptions that deviate from these periodic behaviors. To model these variations, we compute the residual component as the difference between the original traffic data $\mathbf{X}$ and the hybrid periodic pattern $\mathbf{S}^{\mathrm{in}}$:

\begin{equation}
    \mathbf{R}^{\mathrm{in}} = \mathbf{X} - \mathbf{S}^{\mathrm{in}}.
\end{equation}

These residual fluctuations often exhibit high-frequency variations across both space and time. To better capture these dynamics, the Frequency-Aware Residual Representation Module applies a \textbf{Spatial-Temporal Frequency Encoder} (STFE), which transforms the residual component into the frequency domain. This encoder models frequency-specific spatial and temporal behaviors, allowing the network to capture the high-frequency residual variations and complement the periodic patterns for more accurate forecasting.

\subsubsection{Spatial-Temporal Frequency Encoder}
To capture spatial-temporal interactions in the residual component, we transform it into the frequency domain along both the spatial and temporal dimensions. In each dimension, we apply a complex-valued MLP to model frequency-specific behaviors. The transformed residual is then converted back to the time domain. 

We first project the residual component $\mathbf{R}^{\mathrm{in}} \in \mathbb{R}^{T_1 \times N}$ into a high-dimensional space, resulting in $\tilde{\mathbf{R}}^{\mathrm{in}} \in \mathbb{R}^{T_1 \times N \times D}$, where $D$ is the embedding dimension. To enable frequency-domain modeling, we apply Fast Fourier Transform (FFT) along both dimensions: 
\begin{equation}
    \mathcal{R}_f^s = \mathrm{FFT}_s(\tilde{\mathbf{R}}^{\mathrm{in}}, \mathrm{dim}=\mathrm{spatial}),
\end{equation}
where $\mathcal{R}_f^s \in \mathbb{C}^{T_1 \times N' \times D}$ is the spatial frequency representation, and $N' = \left\lfloor \frac{N}{2} \right\rfloor + 1$ corresponds to the truncated spectrum due to symmetry in FFT.

Motivated by recent advances in frequency modeling~\cite{yi2023frequency}, we introduce a complex-valued MLP (C-MLP) to refine the frequency representation. Operating in the frequency domain with C-MLP is equivalent to convolution in the time domain, enabling more efficient and global modeling while reducing computational cost by focusing on the essential frequency components:
\begin{equation}
    \begin{split}
        \mathrm{Re}(\mathcal{R}_1) &= \sigma \left(\mathrm{Re}(\mathcal{R}_f^s) \cdot \mathbf{W}_1^r - \mathrm{Im}(\mathcal{R}_f^s) \cdot \mathbf{W}_1^i + \mathbf{b}_1^r \right) \\
        \mathrm{Im}(\mathcal{R}_1) &= \sigma \left(\mathrm{Im}(\mathcal{R}_f^s) \cdot \mathbf{W}_1^r + \mathrm{Re}(\mathcal{R}_f^s) \cdot \mathbf{W}_1^i + \mathbf{b}_1^i \right) \\
        \mathrm{Re}(\mathcal{R}_2) &= \sigma \left(\mathrm{Re}(\mathcal{R}_1) \cdot \mathbf{W}_2^r - \mathrm{Im}(\mathcal{R}_1) \cdot \mathbf{W}_2^i + \mathbf{b}_2^r \right) \\
        \mathrm{Im}(\mathcal{R}_2) &= \sigma \left(\mathrm{Im}(\mathcal{R}_1) \cdot \mathbf{W}_2^r + \mathrm{Re}(\mathcal{R}_1) \cdot \mathbf{W}_2^i + \mathbf{b}_2^i \right),
    \end{split}
\label{equ:C-MLP}
\end{equation}
where $\mathcal{R}_2 \in \mathbb{C}^{T_1 \times N' \times D}$ is the output after applying C-MLP to the spatial frequency representation, and $\mathrm{Re}(\cdot)$ and $\mathrm{Im}(\cdot)$ denote the real and imaginary parts of a complex-valued tensor, respectively. 
The weights $\mathbf{W}_1^r, \mathbf{W}_1^i \in \mathbb{R}^{D*D'}$, $\mathbf{W}_2^r, \mathbf{W}_2^i \in \mathbb{R}^{D'*D}$ and $b_1^r, b_1^i \in \mathbb{R}^{D'}$, $b_2^r, b_2^i \in \mathbb{R}^D$ are the trainable parameters, and $\sigma$ denotes the ReLU activation function. $D'$ is the hidden layer dimension.

After the frequency domain refinement, we apply the Inverse Fast Fourier Transform (IFFT) to convert the frequency representation back into the time domain:
\begin{equation}
    \mathbf{R}^s = \mathrm{IFFT}_s(\mathcal{R}_2, \mathrm{dim} = \mathrm{spatial)},
\end{equation}
where $\mathbf{R}^s \in \mathbb{R}^{T_1 \times N \times D}$ is the refined residual representation in the spatial domain. Similarly, we repeat the same process (FFT, C-MLP, IFFT) along the temporal dimension to capture the temporal frequency behaviors:
\begin{equation}
    \begin{split}
        \mathcal{R}_f^t &= \mathrm{FFT}_t(\mathbf{R}^{\mathrm{s}}, \mathrm{dim}=\mathrm{temporal}) \\
        \mathcal{R}_2' &= \text{C-MLP}(\mathcal{R}_f^t) \\
        \mathbf{R}^t &= \mathrm{IFFT}_t(\mathcal{R}_2', \mathrm{dim}=\mathrm{temporal}).
    \end{split}
\end{equation}

Finally, we add the residual connection and apply a linear projection to forecast $T_2$ future time steps:
\begin{equation}
    \mathbf{R}^{\mathrm{out}} = \mathrm{Proj}(\mathbf{R}^t + \tilde{\mathbf{R}}^{\mathrm{in}}).
\end{equation}

The final prediction is then computed by combining the outputs of the hybrid periodic component $\mathbf{S}^{\mathrm{out}}$ and the residual component $\mathbf{R}^{\mathrm{out}}$:
\begin{equation}
    \hat{\mathbf{Y}} = \mathbf{S}^{\mathrm{out}} + \mathbf{R}^{\mathrm{out}}.
\end{equation}

After obtaining the final prediction $\hat{\mathbf{Y}}$ by combining the hybrid periodic component $\mathbf{S}^{\mathrm{out}}$ and the residual component $\mathbf{R}^{\mathrm{out}}$, we need to ensure the two components remain distinct to maximize their complementary information. To achieve this, we introduce the Dual-View Alignment Loss (DVA).

\subsection{Dual-View Alignment Loss}
Despite being processed separately, the hybrid periodic and residual components may still overlap in their representations, reducing the decoupling's effectiveness. To address this, the DVA explicitly enforces separation between the low-frequency periodic component and the high-frequency residual component.

We apply FFT to the predicted output $\hat{\mathbf{Y}}$ to obtain its frequency representation ${\hat{\mathcal{Y}}}_f
= \mathrm{FFT}(\hat{\mathbf{Y}})$, and divide the frequency spectrum into low- and high-frequency parts using a predefined threshold $F_{\mathrm{low}}$:
\begin{equation}
\hat{\mathcal{Y}}_f^{\mathrm{low}} = \hat{\mathcal{Y}_f} [0:F_{\mathrm{low}}] \quad
\hat{\mathcal{Y}}_f^{\mathrm{high}} = \hat{\mathcal{Y}_f} [F_{\mathrm{low}}:],
\end{equation}
where $\hat{\mathcal{Y}}_f^{\mathrm{low}}$ represents the low-frequency part, which predominantly captures the hybrid periodic patterns (i.e., the periodic component), and $\hat{\mathcal{Y}}_f^{\mathrm{high}}$ represents the high-frequency part, which captures sharp fluctuations and non-periodic variations (i.e., the residual component).

Next, we apply the IFFT to each part, recovering the corresponding time domain representations. The low frequency part is compared with the hybrid periodic pattern $\mathbf{S}^{\mathrm{out}}$, while the high frequency part is compared with the residual $\mathbf{R}^{\mathrm{out}}$, both using the Mean Squared Errors (MSE):
{
\small
\begin{equation}
    \mathcal{L}_{\mathrm{dva}} = \mathrm{MSE}(\mathrm{IFFT}(\hat{\mathcal{Y}}^{\mathrm{low}}_f), \mathbf{S}^{\mathrm{out}}) + 
    \mathrm{MSE}(\mathrm{IFFT}(\hat{\mathcal{Y}}^{\mathrm{high}}_f), \mathbf{R}^{\mathrm{out}}).
\end{equation}
}


This loss enforces frequency alignment by encouraging the low-frequency part of the prediction to match the periodic component and the high-frequency part to match the residual component, ensuring that the two branches capture distinct and complementary information.

The prediction loss $\mathcal{L}_{\mathrm{pred}}$ is computed as:
\begin{equation}
    \begin{split}
        \mathcal{L}_{\mathrm{pred}} &= \sum_{t=T_1+1}^{T_1+T_2} \sum_{n=1}^{N} \left| \hat{\mathbf{Y}}_{t,n} - \mathbf{Y}_{t,n}^{gt} \right|.
    \end{split}
\end{equation}
where $\hat{\mathbf{Y}}_{t,n}$ denotes the predicted value at time step $t$ for node $n$, and $\mathbf{Y}_{t,n}^{gt}$ is the corresponding ground-truth.

Finally, the total training loss $\mathcal{L}$ combines the prediction loss with the alignment loss:
\begin{equation}
    \begin{split}
        \mathcal{L} &= \mathcal{L}_{\mathrm{pred}} + \alpha *  \mathcal{L}_{\mathrm{dva}},
    \end{split}
\end{equation}
where $\alpha$ is a weighting coefficient that balances the contribution of the alignment loss.
\section{Experiments}
In this section, we conduct comprehensive experiments to evaluate \name from multiple perspectives. We aim to answer the following key research questions:

\textbf{Q1:} How does \name perform compared to SOTA prediction methods across diverse real-world traffic datasets?

\textbf{Q2:} What is the impact of each module on the overall performance of \name?

\textbf{Q3:} How does \name perform in terms of reliability and computational performance under real-world disturbances?


\subsection{Datasets}
To assess the performance of \name, we conduct experiments on four commonly used real-world traffic flow datasets: PEMS03/04/07/08~\cite{song2020spatial}.

\subsection{Baselines}
To comprehensively evaluate our \name, we compare with two lines of state-of-the-art methods, including \textbf{(a) spatial-temporal prediction methods}: STGCN~\cite{yu2018spatio}, DCRNN~\cite{li2018diffusion}, GWNet~\cite{wu2019graph}, ASTGCN~\cite{guo2019attention}, MTGNN~\cite{wu2020connecting}, STGODE~\cite{fang2021spatial}, ST-WA~\cite{cirstea2022towards}, DGCRN~\cite{li2023dynamic} and STPGNN~\cite{kong2024spatio}, and \textbf{(b) spatial-temporal decoupling methods}: D\textsuperscript{2}STGNN~\cite{shao2022decoupled}, STWave~\cite{fang2023spatio}, CycleNet~\cite{cyclenet}, and STDN~\cite{cao2025spatiotemporal}. 
We compare two CycleNet variants: CycleNet-W and CycleNet-D, both built on an MLP backbone. 

\subsection{Experimental Setup}
To ensure a fair comparison, we adopt the experimental settings commonly used in previous studies~\cite{shao2024exploring}. The datasets are divided into training, validation, and test sets with a ratio of 6:2:2, respectively. We use the past 12 time steps (previous hour) to forecast the next 12 time steps (upcoming hour). 


\begin{table*}[ht!]
\renewcommand{\arraystretch}{1.3} 
\centering
\scalebox{0.8}{
\begin{tabular}{c|l|ccc|ccc|ccc|ccc}
\toprule
\multirow{2}{*}{Method} & Dataset & \multicolumn{3}{c|}{PEMS03} & \multicolumn{3}{c|}{PEMS04} & \multicolumn{3}{c|}{PEMS07} & \multicolumn{3}{c}{PEMS08} \\ 
\cmidrule(lr){2-2} \cmidrule(lr){3-5} \cmidrule(lr){6-8} \cmidrule(lr){9-11} \cmidrule(lr){12-14}
& Metric & MAE & RMSE & MAPE & MAE & RMSE & MAPE & MAE & RMSE & MAPE & MAE & RMSE & MAPE \\ \midrule
\multirow{9}{*}{\makecell[c]{Spatial-Temporal\\ Prediction Methods}} & STGCN & 17.49 & 30.12 & 17.15\% & 22.70 & 35.55 & 14.59\% & 25.38 & 38.78 & 11.08\% & 18.02 & 27.83 & 11.40\% \\
& DCRNN & 18.18 & 30.31 & 18.91\% & 24.70 & 38.12 & 17.12\% & 25.30 & 38.58 & 11.66\% & 17.86 & 27.83 & 11.45\% \\ 
& GWNet & 19.85 & 32.94 & 19.31\% & 25.45 & 39.70 & 17.29\% & 26.85 & 42.78 & 12.12\% & 19.13 & 31.05 & 12.68\% \\ 
& ASTGCN & 17.69 & 29.66 & 19.40\% & 22.93 & 35.22 & 16.56\% & 28.05 & 42.57 & 13.92\% & 18.61 & 28.16 & 13.08\% \\ 
& MTGNN & 17.23 & 25.89 & 17.35\% & 19.98 & 31.92 & 14.13\% & 23.92 & 35.86 & 12.43\% & 15.03 & 23.89 & 10.23\% \\
& STGODE & 16.50 & 27.84 & 16.69\% & 20.84 & 32.82 & 13.77\% & 22.99 & 37.54 & 10.14\% & 16.81 & 25.97 & 10.62\% \\ 
& ST-WA & 15.17 & 26.63 & 15.83\% & 19.06 & 31.02 & 12.52\% & 20.74 & 34.05 & 8.77\% & 15.41 & 24.62 & 9.94\%  \\ 
& DGCRN & \underline{14.74} & 25.97 & \underline{15.42}\% & 18.80 & 30.65 & 12.82\% & 20.48 & 33.25 & 9.06\% & 14.60 & 24.16 & 9.33\%  \\
& STPGNN & 14.87 & \underline{25.89} & 15.54\% & 18.86 & 30.13 & 13.14\% & 21.77 & 35.28 & 9.37\% & 14.69 & 23.85 & 9.55\% \\ \midrule
\multirow{6}{*}{\makecell[c]{Spatial-Temporal \\ Decoupling Methods}} & D\textsuperscript{2}STGNN & 15.10 & 26.57 & \textbf{15.23}\% & \underline{18.42} & \underline{29.97} & 12.81\% & \underline{19.68} & \underline{33.24} & 8.43\% & 14.35 & 24.18 & 9.33\%  \\ 
& STWave & 14.84 & 26.20 & 15.86\% & 18.57 & 30.24 & \underline{12.57}\% & 19.72 & 33.72 & \underline{8.19}\% & \underline{13.84} & \underline{23.60} & \underline{9.19}\%  \\ 
& CycleNet-D & 17.54 & 27.48 & 20.02\% & 23.34 & 36.11 & 16.94\% & 25.55 & 39.82 & 11.82\% & 18.89 & 29.20 & 13.42\% \\
& CycleNet-W & 18.34 & 28.57 & 20.25\% & 24.29 & 38.13 & 18.00\% & 24.99 & 40.25 & 11.01\% & 18.27 & 29.83 & 11.89\% \\
& STDN & 16.05 & 27.51 & 17.71\% & 18.67 & 30.92 & 13.16\% & 22.94 & 36.06 & 10.32\% & 14.79 & 24.60 & 10.26\% \\ 
\cmidrule(lr){2-14}
& \textbf{\name} & \textbf{14.69} & \textbf{23.66} & 15.76\% & \textbf{18.20} & \textbf{29.94} & \textbf{12.44}\% & \textbf{19.37} & \textbf{33.22} & \textbf{8.05}\% & \textbf{13.59} & \textbf{23.24} & \textbf{8.91}\%  \\ \bottomrule
\end{tabular}}
\caption{Performance comparison of all models on four real-world datasets. The best results are highlighted in \textbf{bold}, while the second-best results are \underline{underlined}. All results are the average value of 5 repetitions. 
}
\label{tab:main_results}
\end{table*}

\subsection{Main Results}
To answer Q1, we evaluate the overall prediction performance of \name against existing methods, as shown in Table~\ref{tab:main_results}. \name outperforms all compared models across the datasets, demonstrating the effectiveness of our approach. Key findings are as follows:

\ding{192} \textbf{Decoupled methods generally outperform conventional spatial-temporal prediction methods}. While models like D\textsuperscript{2}STGNN and STWave perform suboptimally on three out of four datasets, their strong performance can be attributed to their decoupled architectures, which separate trend and seasonal components for more focused modeling. Among non-decoupled methods, DGCRN performs well, leveraging graph convolutions to capture spatial dependencies and recurrent layers for temporal dependencies.

\ding{193} \textbf{\name outperforms all decoupled methods, showcasing the advantage of explicitly modeling periodic patterns.} \name reduces average MAE by 22.63\% and 23.27\% compared to CycleNet-D and CycleNet-W, respectively. This highlights that \name’s hybrid periodic pattern, aided by spatial-temporal modeling, provides a more comprehensive representation of multi-scale periodic patterns compared to the single-scale design in CycleNet.




\begin{table}[t!]
\small
\setlength{\tabcolsep}{3.5pt}
\centering
\begin{tabular}{cccc|ccc}
\toprule
Dataset & \multicolumn{3}{c|}{PEMS04} & \multicolumn{3}{c}{PEMS08} \\
\cmidrule(lr){1-1} \cmidrule(lr){2-4} \cmidrule(lr){5-7}
Metric & MAE & RMSE & MAPE & MAE & RMSE & MAPE \\
\midrule
w/o LDE & 18.30 & 30.02 & 12.59\% & 13.64 & 23.37 & 9.01\% \\
w/o LWE & 19.80 & 31.37 & 13.92\% & 15.83 & 25.05 & 10.43\% \\ \midrule
w/o STAE & 21.07 & 33.46 & 15.82\% & 15.97 & 25.96 & 11.12\% \\
w/o STFE & 25.24 & 41.07 & 17.57\% & 18.47 & 31.27 & 12.12\% \\
Re-MLP & 18.57 & 30.52 & 12.75\% & 13.73 & 23.42 & 9.07\% \\ \midrule
w/o DVA & 18.47 & 30.16 & 12.61\% & 13.70 & 23.44 & 9.08\% \\ \midrule
\textbf{\name} & \textbf{18.20} & \textbf{29.94} & \textbf{12.44\%} & \textbf{13.59} & \textbf{23.24} & \textbf{8.91\%} \\
\bottomrule
\end{tabular}
\caption{Ablation study results in the PEMS04 and PEMS08 datasets.}
\label{tab:ablation}
\end{table}

\subsection{Ablation Study}
To answer Q2, we conduct ablation studies in the PEMS04 and PEMS08 datasets to assess the contribution of each module in \name.

\ding{194} \textbf{Each module in \name significantly contributes to its overall performance.} Removing any module leads to a noticeable decrease in accuracy, highlighting the importance of each component. As shown in Table~\ref{tab:ablation}, we evaluate several variants: \textbf{w/o LDE}, \textbf{w/o LWE}, \textbf{w/o STAE}, \textbf{w/o STFE}, \textbf{Re-MLP} (replacing STFE with a simple MLP), and \textbf{w/o DVA}.

Notably, removing the LWE leads to greater degradation than LDE. For example, in the PEMS04 dataset, the MAE rises from 18.20 to 19.80 without LWE, whereas it increases only to 18.30 without LDE. This is because weekly patterns capture longer periodic structures, including both weekday and weekend dynamics, thereby providing richer information. 
Additionally, removing either STFE or STAE significantly degrades performance. For instance, in the PEMS04 dataset, removing the STAE increases the MAE from 18.20 to 21.07, while removing the STFE leads to an even larger increase to 25.24. This suggests that the spatial-temporal modeling of the residual component plays a more critical role than that of the periodic patterns, which captures sudden and irregular fluctuations that are not explained by regular periodic patterns.

\begin{table}[t!]
\small
\setlength{\tabcolsep}{4pt}
\centering
\begin{tabular}{cccc|ccc}
\toprule
Dataset & \multicolumn{3}{c|}{PEMS04} & \multicolumn{3}{c}{PEMS08} \\
\cmidrule(lr){1-1} \cmidrule(lr){2-4} \cmidrule(lr){5-7}
Metric & MAE & RMSE & MAPE & MAE & RMSE & MAPE \\
\midrule
Zero & 18.49 & 30.22 & 12.79\% & 14.21 & 23.90 & 9.64\% \\
Uniform & 18.48 & 30.28 & 12.67\% & 13.80 & 23.45 & 9.19\% \\
Normal & 18.76 & 30.48 & 13.17\% & 13.89 & 23.59 & 9.26\% \\
Xavier & 18.43 & 30.20 & 12.72\% & 13.82 & 23.47 & 9.22\% \\
He & 18.45 & 30.19 & 12.70\% & 13.74 & 23.33 & 9.19\% \\
\midrule
\textbf{\name} & \textbf{18.20} & \textbf{29.94} & \textbf{12.44\%} & \textbf{13.59} & \textbf{23.24} & \textbf{8.91\%} \\
\bottomrule
\end{tabular}
\caption{Performance comparison of different initialization strategies in the PEMS04 and PEMS08 datasets.}
\label{tab:ablation_init}
\end{table}

\ding{195} \textbf{Our statistical prior initialization outperforms all other initialization strategies, effectively learning meaningful periodic embeddings.}  As shown in Table~\ref{tab:ablation_init}, \name consistently achieves the best performance across both the PEMS04 and PEMS08 datasets. This superior performance, reflected in the lower MAE, RMSE, and MAPE values, demonstrates that statistical prior initialization significantly enhances model convergence and the ability to learn accurate periodic patterns. Compared to other initialization strategies, \name ensures faster and more stable training, leading to better overall results. Detailed descriptions of these initialization strategies are provided in Appendix C.4.

\begin{figure}[t!]
  \centering
  \includegraphics[width=0.49\textwidth]{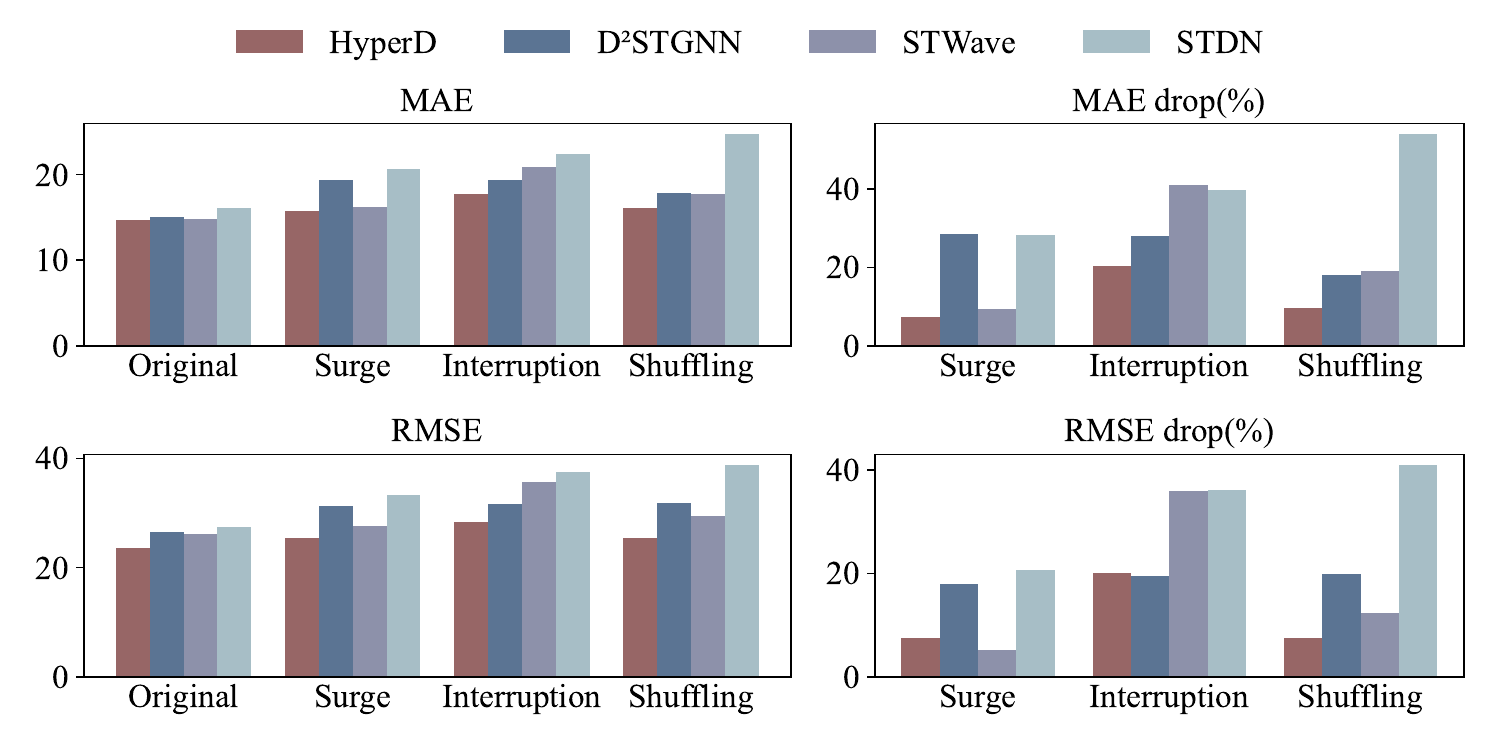} 
  \caption{Robustness testing in the PEMS03 datasets.}
  \label{fig:robustness} 
\end{figure}

\subsection{Reliability and Computational Performance}
To answer Q3, we evaluate the reliability and computational performance of \name under real-world disturbances and measure its efficiency in terms of computational time and memory consumption.

\ding{196}  \textbf{\name outperforms all other models, maintaining the lowest performance drop under various traffic perturbations, highlighting its superior robustness.} We simulate three types of traffic perturbations to evaluate the robustness of \name: (1) \textbf{Sudden surge } (values ×1.5), (2) \textbf{Sudden interruption} (values set to zero), (3) \textbf{Segment shuffling} (values over four time steps reordered). 
Results for PEMS03 are shown in Figure~\ref{fig:robustness}, and additional results and experimental settings are provided in Appendix C.5.
While all models suffer performance degradation under these disturbances, \name consistently outperforms the other decoupled models, achieving the lowest MAE and RMSE across all types of perturbations. Notably, \name exhibits the smallest relative performance drop, indicating its robustness in maintaining performance even when faced with different traffic disruptions. In contrast, D\textsuperscript{2}STGNN and STDN show significant performance declines, particularly under the sudden interruption perturbation. Although STWave performs better than the others, it still falls short of \name in handling disruptions, confirming the advantage of our approach in handling real-world traffic disturbances.

\begin{table}[t!]
\small
\centering
\scalebox{0.9}{
\begin{tabular}{lccc}
\toprule
Model & Max Mem.(GB) & Epoch Time(s) & Infer Time(CPU) \\ \midrule
D\textsuperscript{2}STGNN & 17.90 & 57.29 & 8.00 \\
STWave & \underline{10.28} & \underline{49.65} & 7.11 \\
STDN & 15.65 & 59.20 & \underline{7.10} \\ \midrule
\textbf{\name} & \textbf{1.64} & \textbf{7.31} & \textbf{1.49} \\
\textit{Reduction} & $\textit{6.27}\times\!\downarrow$ & $\textit{6.79}\times\!\downarrow$ & $\textit{4.77}\times\!\downarrow$ \\
\bottomrule
\end{tabular}}
\caption{Efficiency comparison of \name and other decoupled models in the PEMS04 dataset with a batch size of 64.}
\label{tab:efficiency_comparison}
\end{table}

\begin{figure}[ht!]
  \centering
  \includegraphics[width=0.49\textwidth]{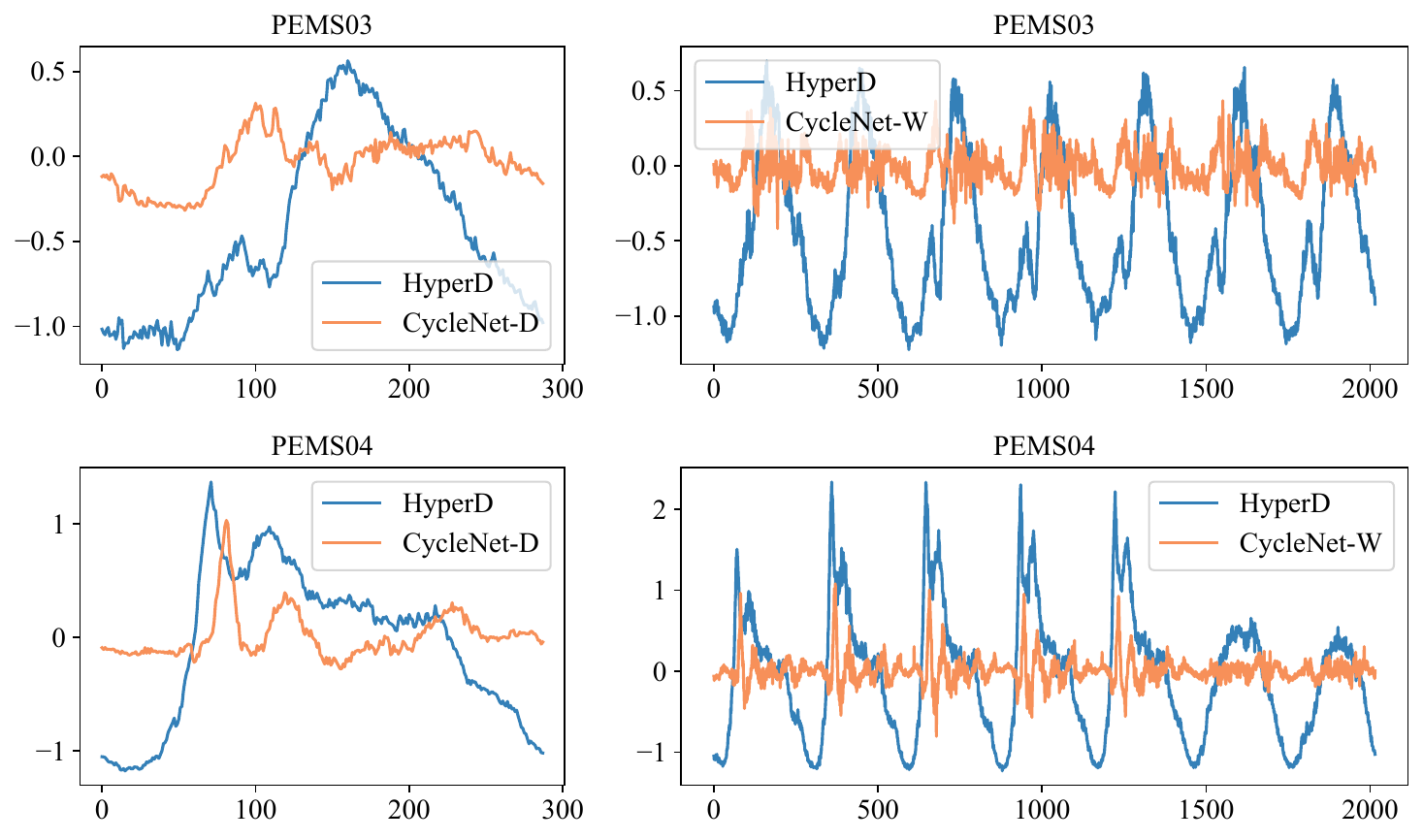} 
  \caption{Visualization of the daily and weekly embeddings learned by \name, as well as the daily embedding from CycleNet-D and the weekly embedding from CycleNet-W, in the PEMS03 and PEMS04 datasets.}
  \label{fig:visualization} 
\end{figure}

\ding{197} \textbf{\name demonstrates clear advantages in memory usage, training speed, and inference time, confirming its efficiency for real-time applications.} We assess the computational performance of \name by comparing it with other decoupled models on the PEMS04 dataset, focusing on three key metrics: \textbf{Maximum Memory}(peak GPU usage), \textbf{Epoch Time}(average training time per epoch), \textbf{Infer Time}(average prediction time per instance). As presented in Table~\ref{tab:efficiency_comparison}, \name demonstrates clear advantages in computational efficiency. It requires significantly less GPU memory than other models, trains faster per epoch, and achieves lower inference latency on the CPU. These improvements reflect the lightweight design of \name and the effectiveness of its architectural simplifications, making it well-suited for real-world and resource-constrained applications.

\subsection{Visualization of Periodic Embeddings}

Although not central to our main research questions, we also investigate whether \name can learn more expressive periodic representations compared to prior methods. 
\ding{198}  \textbf{\name captures more expressive and detailed periodic patterns, setting a new standard in modeling temporal dynamics and outperforming existing methods.}
As shown in Figure~\ref{fig:visualization}, \name successfully learns detailed daily and weekly embeddings, whereas CycleNet-D and CycleNet-W yield compressed ones. The embeddings learned by \name are more stretched and expanded along the temporal axis, capturing clearer and more detailed periodic patterns. 
This superior expressiveness, enabling clearer periodic patterns, stems from \name's spatial-temporal modeling that facilitates long-range interactions.

\section{Conclusion}
In this paper, we introduce \name, a novel Hybrid Periodicity Decoupling Framework for traffic forecasting. \name decouples traffic data into periodic and residual components, each processed by the Hybrid Periodic Representation Module and the Frequency-Aware Residual Representation Module, respectively. We further introduce a Dual-View Alignment Loss to promote effective and thorough decoupling between the two components. Extensive experiments on four real-world datasets demonstrate that \name surpasses state-of-the-art methods in forecasting accuracy, robustness, and computational efficiency, highlighting the importance of explicitly modeling periodic patterns.

\section{Acknowledgment}
This work was supported by a grant from the National Natural Science Foundation of China under grants (No. 62372211, 62272191), the Science and Technology Development Program of Jilin Province (No. 20250102216JC), and the China Postdoctoral Science Foundation under Grant Number (2025M771587).

\bibliography{references}






\appendix
\section{Architectural Details}
\begin{figure}[h]
    \centering
    \begin{subcaptionbox}{Spatial-Temporal Attentive Encoder. \label{fig:STAE}}[.45\linewidth]
        {\includegraphics[height=3.5cm]{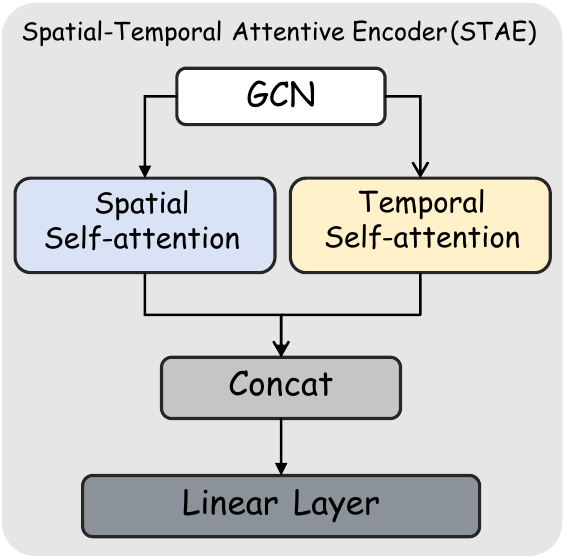}}
    \end{subcaptionbox}
    \hspace{0.01\linewidth}  
    \begin{subcaptionbox}{Spatial-Temporal Frequency Encoder. \label{fig:STFE}}[.45\linewidth]
        {\includegraphics[height=3.5cm]{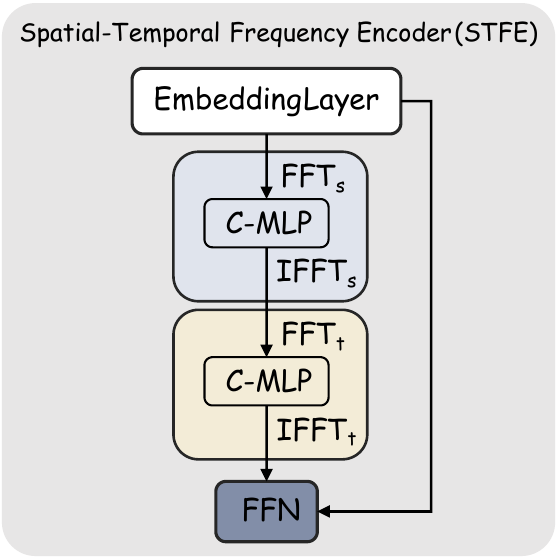}}
    \end{subcaptionbox}
    \caption{Detailed architectures of the STAE and the STFE.}
    \label{fig:encoders}
\end{figure}

In this section, we introduce the detailed architectures of the Spatial-Temporal Attentive Encoder (STAE) and the Spatial-Temporal Frequency Encoder (STFE) used in the Hybrid Periodic Representation Module and the Frequency-Aware Residual Representation Module, respectively. 

As shown in Figure~\ref{fig:STAE}, the STAE takes the learnable periodic embedding as input and is designed to model the spatial and temporal dependencies among periodic patterns. It consists of a Graph Convolutional Network (GCN), a spatial self-attention mechanism, and a temporal self-attention mechanism. The GCN is employed to capture localized spatial correlations by aggregating information from neighboring nodes. The spatial and temporal self-attention mechanisms are used in parallel to model long-range dependencies along their respective dimensions. The outputs of the two self-attention branches are concatenated and then passed through a linear layer to generate the refined embedding. Through this design, the STAE effectively models both local and global dependencies in the periodic embedding.

As shown in Figure~\ref{fig:STFE}, the STFE takes the residual component as input and is designed to extract frequency-aware representations from both spatial and temporal perspectives. The input is first projected into a high-dimensional space through an embedding layer. It then undergoes a sequence of operations in frequency domain along the spatial and temporal dimensions. Specifically, the input is first transformed into the frequency domain via the FFT, then passed through a complex-valued MLP (C-MLP) for representation learning, and finally mapped back to the time domain via the IFFT. This process is applied sequentially along both the spatial and temporal dimensions, allowing the encoder to capture global irregular fluctuations that are otherwise difficult to learn in the time domain.

\section{Preliminaries of the Fast Fourier Transform (FFT)}

In this section, we introduce the basic concepts of the Discrete Fourier Transform (DFT) and its efficient implementation via the Fast Fourier Transform (FFT). 
Formally, for a discrete time series  $x = \{ x_t \}_{t=0}^{T-1}$, the DFT transforms it into the frequency domain by projecting it onto orthogonal sinusoidal bases:
\begin{equation}
    X_f(k) = \sum_{t=0}^{T-1} x_t \cdot e^{-j2\pi k t/T}, \quad k = 0, 1, \ldots, T - 1,
\end{equation}
where $X_f(k)$ denotes the complex frequency coefficient in the frequency index $k$, and $j$ is the imaginary unit. 

In practical applications, directly computing the DFT has a computational complexity of $O(T^2)$, which becomes prohibitive for long sequences. To address this, we utilize the FFT~\cite{duhamel1990fast}, an efficient algorithm that reduces the complexity to $O(T\log T)$ by exploiting the symmetry properties of the complex exponential terms. The IFFT allows efficient transformation back to the time domain, and is defined as:
\begin{equation}
    x_t = \frac{1}{T} \sum_{k=0}^{T-1} X_f(k) \cdot e^{j2\pi k t/T}, \quad t = 0, 1, \ldots, T - 1.
\end{equation}

These formulations provide the theoretical foundation for the frequency-domain operations employed in our model.

\section{Experimental Details}

\subsection{Datasets}

\begin{table}[t!]
\setlength{\tabcolsep}{3.5pt}
\centering
\scalebox{0.9}{
\begin{tabular}{lccc}
\toprule
Dataset & \#Sensors (N) & \#Timesteps & Time Range \\
\midrule
PEMS03 & 358 & 26,208 & 09/01/2018 -- 11/30/2018 \\
PEMS04 & 307 & 16,992 & 01/01/2018 -- 02/28/2018 \\
PEMS07 & 883 & 28,224 & 05/01/2017 -- 08/06/2017 \\
PEMS08 & 170 & 17,856 & 07/01/2016 -- 08/31/2016 \\
\bottomrule
\end{tabular}
}
\caption{Statistics of the Datasets.}
\label{tab:dataset}
\end{table}

We adopt four real-world commonly used traffic flow datasets, with the basic statistics summarized in Table~\ref{tab:dataset}.

\subsection{Baselines}
We adopt fourteen representative and state-of-the-art baselines for comparison including spatial-temporal prediction methods and spatial-temporal decoupling methods. We introduce these models as follows:

\textbf{(1) Spatial-Temporal Prediction Methods}

\textbf{STGCN}~\cite{yu2018spatio} integrates graph convolutions for spatial modeling and gated temporal convolutions to capture sequential dynamics over time.

\textbf{DCRNN}~\cite{li2018diffusion} leverages bidirectional diffusion processes to capture spatial correlations, while GRU cells model temporal dependencies.

\textbf{GWNet}~\cite{wu2019graph} combines dilated temporal convolutions with adaptive graph convolution to model dynamic temporal patterns and flexible spatial dependencies.

\textbf{ASTGCN}~\cite{guo2019attention} introduces spatial-temporal attention mechanisms on top of graph convolution to dynamically weigh the importance of nodes and time steps.

\textbf{MTGNN}~\cite{wu2020connecting} jointly learns graph structures and spatial-temporal dependencies using mix-hop graph propagation and dilated temporal convolutions for multivariate time series forecasting.

\textbf{STGODE}~\cite{fang2021spatial} models continuous-time traffic dynamics by integrating Graph Neural ODEs with temporal evolution, enabling the modeling of traffic data as a continuous process rather than in discrete steps.

\textbf{ST-WA}~\cite{cirstea2022towards} generates dynamic, location-aware model parameters from stochastic spatial-temporal representations and incorporates a window attention mechanism to enable efficient spatial-temporal aware learning.

\textbf{DGCRN}~\cite{li2023dynamic} employs hyper-networks to dynamically generate graph structures from node attributes at each time step, combining them with a static graph to capture fine-grained spatial-temporal dynamics.

\textbf{STPGNN}~\cite{kong2024spatio} identifies pivotal nodes through a pivotal node identification module and employs a parallel framework to extract spatial-temporal features from both pivotal and non-pivotal nodes.

\textbf{(2) Spatial-Temporal Decoupling Methods}

\textbf{D\textsuperscript{2}STGNN}~\cite{shao2022decoupled} decomposes traffic flow into diffusion and inherent components for separate modeling and introduces a dynamic graph learning module to capture time-varying spatial dependencies.

\textbf{STWave}~\cite{fang2023spatio} leverages wavelet transforms to decouple traffic signals into stable trends and fluctuating events, which are modeled separately by a dual-branch spatial-temporal network.

\textbf{CycleNet-D} is a re-implementation of CycleNet~\cite{cyclenet} with daily cycle. CycleNet introduces Residual Cycle Forecasting (RCF), which models learnable periodic patterns based on the maximum stable cycle and predicts residuals for long-term time series forecasting.

\textbf{CycleNet-W} is the original model of CycleNet using weekly cycle. 

\textbf{STDN}~\cite{cao2025spatiotemporal} builds a dynamic graph with spatial-temporal embeddings and employs trend-seasonality decomposition mechanism for traffic forecasting.

\subsection{Implementation Details}
All experiments are conducted on a server with NVIDIA L40 GPU cards. The implementation is based on Python 3.12.7 and PyTorch 2.5.1. We use the Adam optimizer and set the batch size to 64 across all experiments. Model performance is evaluated using three standard metrics: Mean Absolute Error (MAE), Root Mean Square Error (RMSE), and Mean Absolute Percentage Error (MAPE). 

\subsection{Ablation Study on Initialization}
To evaluate the effectiveness of our proposed statistical prior initialization strategy, we conduct comparative experiments against five commonly used initialization strategies, which are briefly introduced below:

\textbf{Zero Initialization:} All weights are initialized to zero. While simple, this approach leads to symmetry in parameter updates, preventing effective learning.

\textbf{Uniform Distribution Initialization:} Weights are sampled from a uniform distribution within a fixed range (typically $[-a, a]$), where $a$ is determined based on the layer’s fan-in or fan-out to maintain stable gradients.

\textbf{Normal Distribution Initialization:} Weights are drawn from a normal distribution, often with zero mean and a small standard deviation, assuming no prior structural bias.

\textbf{Xavier Initialization~\cite{glorot2010understanding}:} Also known as Glorot initialization, it sets the weights using a uniform or normal distribution scaled by the number of input and output units, aiming to keep the variance of activations constant across layers.

\textbf{He Initialization~\cite{he2015delving}:} Also known as Kaiming initialization, designed for ReLU activation functions, this method initializes weights from a normal distribution scaled by the number of input units, allowing better signal propagation in deep networks.

All initialization strategies perform worse than our statistical prior initialization. Notably, zero initialization and normal distribution initialization lead to the most significant performance degradation. This suggests that poor initialization can hinder convergence and limit the model’s ability to learn clear periodic patterns.

Our statistical prior initialization leverages domain knowledge by computing the average traffic flow for each node at each time step within the day and week  over the training set. These averages are used to initialize two learnable periodic embeddings, effectively injecting meaningful temporal priors into the model from the outset. This warm-start strategy not only facilitates faster convergence but also provides a strong inductive bias that enhances the model’s ability to capture periodic dynamics.

\subsection{Robustness Testing}

\begin{figure}[t!]
    \centering
    \includegraphics[width=\linewidth]{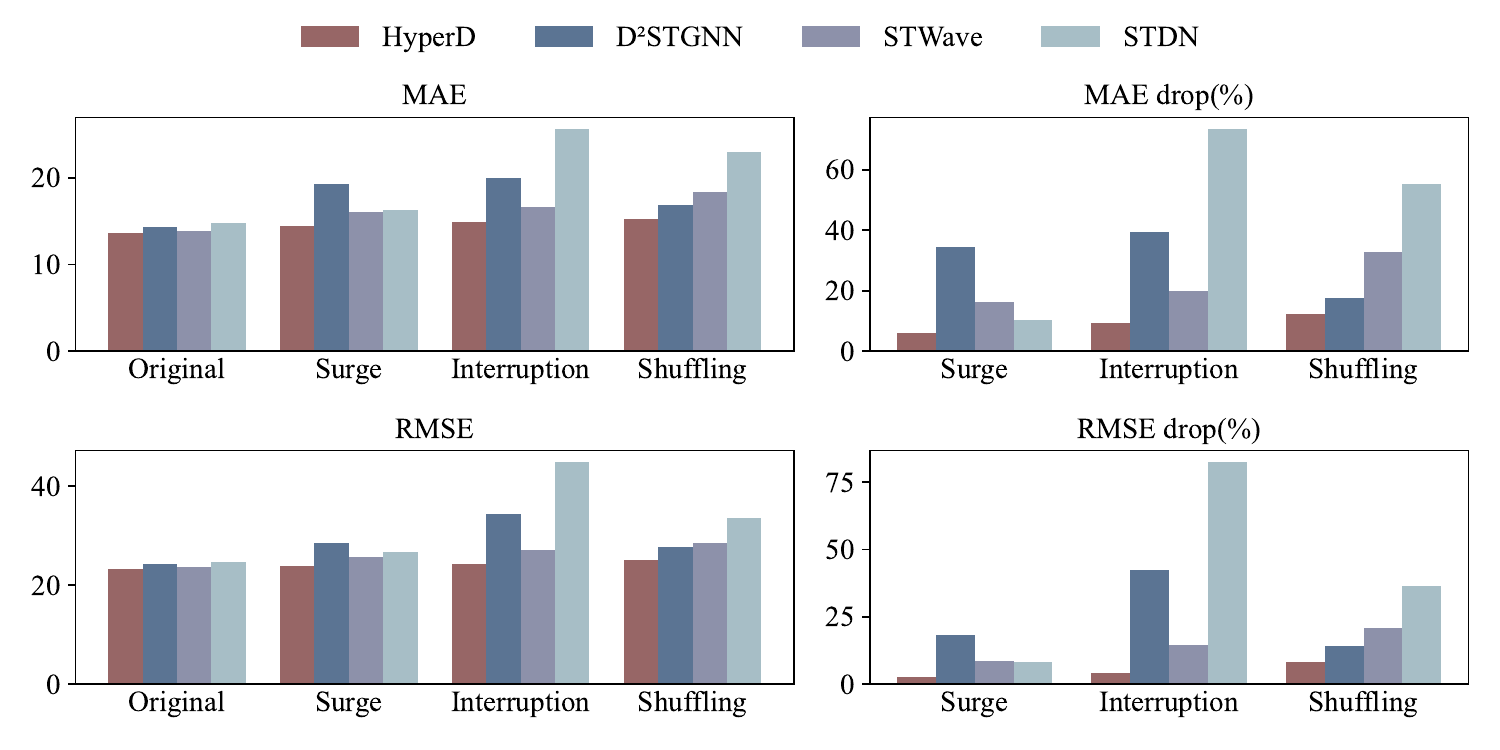}
    \caption{Robustness testing in the PEMS08 datasets.}
    \label{fig:robustness_2}
\end{figure}

Traffic flow prediction often encounters unexpected real-world disturbances, such as sensor failures or sudden traffic surges, which can significantly impact model performance. To evaluate robustness under such conditions, we design three disturbance scenarios applied at a randomly selected time point across all nodes:

\textbf{Sudden surge:} Traffic flow values are multiplied by 1.5 to simulate abrupt traffic increases.

\textbf{Sudden interruption:} Traffic flow values are set to zero, simulating sensor failure or data loss~\cite{yu2025merlin}.

\textbf{Segment shuffling:} Traffic flow values over four future time steps are shuffled to mimic pattern disruption.

The results in the PEMS08 dataset are shown in Figure~\ref{fig:robustness_2}. 
The performance of all models deteriorates under the three designed disturbance scenarios. Among them, HyperD consistently achieves the best MAE and
RMSE under all perturbations. More importantly, HyperD exhibits the smallest percentage of performance dropouts, indicating its strong robustness. The percentage of performance drop reflects the relative degradation compared to the original setting, and a lower value indicates that the model is less sensitive to the injected disturbances. In addition, HyperD shows a balanced robustness across all three scenarios, demonstrating that it can robustly handle diverse disruption patterns.

We attribute this robustness to decoupling the data into the periodic and residual components and the modeling of spatial-temporal dependencies in both components. In contrast, models like D\textsuperscript{2} STGNN and STDN, which lack structural decoupling and explicit modeling of periodic patterns, struggle to adapt when the input deviates significantly from training distributions.

\subsection{Hyper-Parameter Study}

\begin{figure}[t!]
  \centering
  \includegraphics[width=0.49\textwidth]{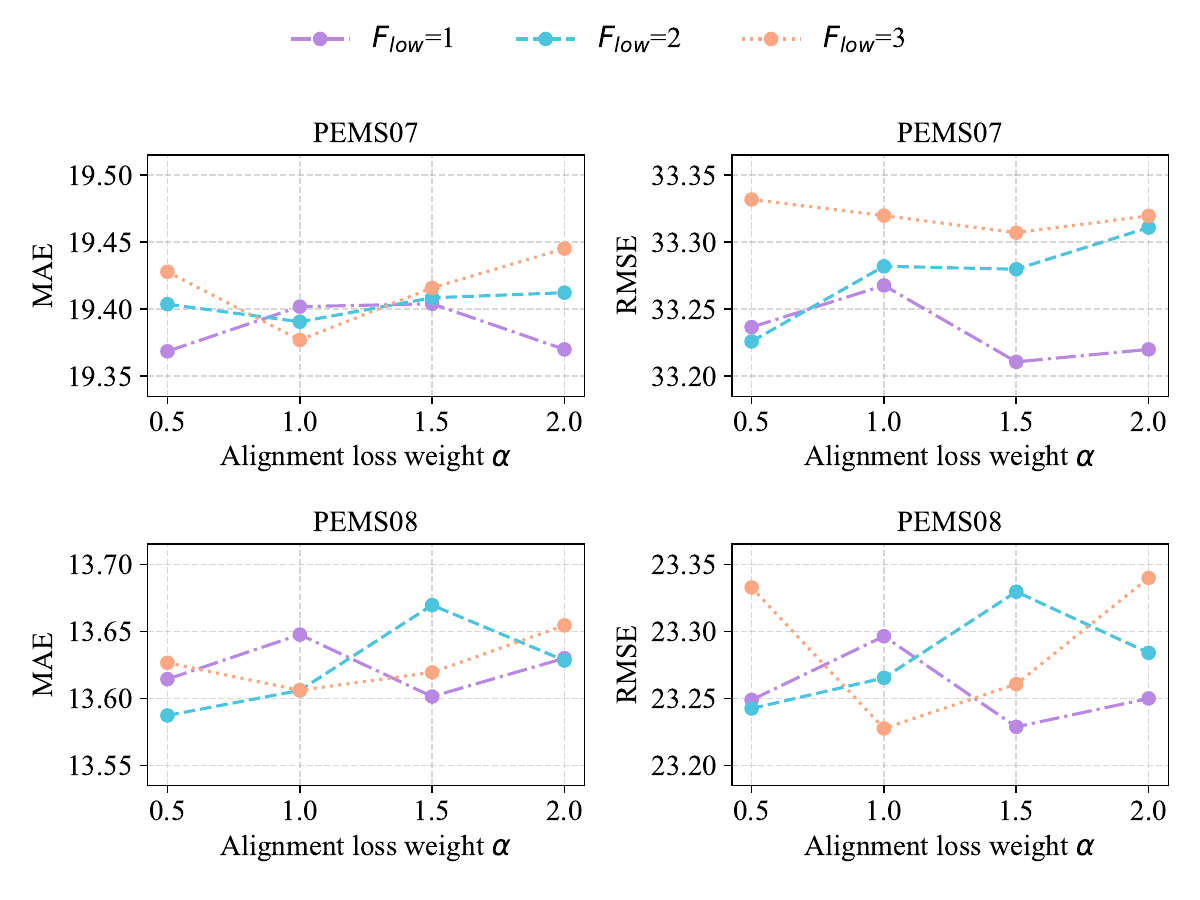} 
  \caption{Hyperparameter study in the PEMS07 and PEMS08 datasets.}
  \label{fig:hyper-param} 
\end{figure}

We conduct experiments on the sensitivity of HyperD to key hyper-parameters, including the cutoff frequency index $F_{\text{low}}$ and the weighting coefficient $\alpha$ in the dual-view alignment loss.
Specifically, we search $F_{\mathrm{low}}$ over the set $\{1, 2, 3\}$ and select $\alpha$ from $\{0.5, 1, 1.5, 2\}$. Figure~\ref{fig:hyper-param} plots the average MAE and RMSE of predictions in the PEMS07 and PEMS08 dataset. We can draw the following conclusions: (1) The optimal values of $F_{\text{low}}$ and $\alpha$ vary across datasets. Specifically, the best performance in the PEMS07 dataset is achieved with $F_{\text{low}}=1$ and $\alpha=2.0$, while the PEMS08 dataset performs best with $F_{\text{low}}=2$ and $\alpha=0.5$. (2) The impact of $F_{\text{low}}$ and $\alpha$ differs between the MAE and RMSE. Overall, the hyper-parameters have a relatively minor effect on the MAE, but a more pronounced influence on the RMSE. The stability of the MAE across different settings indicates that the model is less sensitive to the hyper-parameter changes in terms of average error, suggesting a certain degree of robustness to fluctuations in prediction accuracy.


\end{document}